\documentclass[letterpaper, 12 pt, journal]{ieeeconf}  

\usepackage[english, brazil]{babel}
\usepackage{graphicx}
\usepackage[utf8]{inputenc}

\IEEEoverridecommandlockouts                              
\title{Avaliação de Classificadores para Segmentação de Imagens:\\Aplicações para Inventário Florestal de Eucalipto\thanks{\noindent $\star$ Este trabalho é parte dos requisitos para aprovação na disciplina de Trabalho de Conclusão de Curso (2016/2) do discente Rodrigo M. Ferreira, no curso de Bacharelado em Sistemas de Informação.}}

\author{Rodrigo M. Ferreira e Ricardo M. Marcacini
\\~
\\Universidade Federal de Mato Grosso do Sul
\\Sistemas de Informação
\\Câmpus de Três Lagoas - MS}

\begin{document}

\maketitle
\thispagestyle{empty}
\pagestyle{empty}

\begin{abstract}
The task of counting eucalyptus trees from aerial images collected by unmanned aerial vehicles (UAVs) has been frequently explored by techniques of estimation of the basal area, i.e, by determining the expected number of trees based on sampling techniques. An alternative is the use of machine learning to identify patterns that represent a tree unit, and then search for the occurrence of these patterns throughout the image. This strategy depends on a supervised image segmentation step to define predefined interest regions. Thus, it is possible to automate the counting of eucalyptus trees in these images, thereby increasing the efficiency of the eucalyptus forest inventory management. In this paper, we evaluated 20 different classifiers for the image segmentation task. A real sample was used to analyze the counting trees task considering a practical environment. The results show that it possible to automate this task with 0.7\% counting error, in particular, by using strategies based on a combination of classifiers. Moreover, we present some performance considerations about each classifier that can be useful as a basis for decision-making in future tasks.\\
-----\\
\textit{Resumo} ----- A tarefa de contagem de árvores (ou mudas) de eucalipto por meio de imagens aéreas coletadas por Veículos Aéreos não Tripulados (VANTs) tem sido frequentemente explorado por meio de técnicas de estimativa da área basal, ou seja, que determinam o número esperado de árvores com base em amostragem. Uma estratégia alternativa é o uso de aprendizado de máquina para identificar padrões que representam uma unidade arbórea (copa de árvore ou muda), e verificar a ocorrência desses padrões em toda a imagem. Esta estratégia depende de uma etapa de segmentação supervisionada da imagem, que é o processo para identificar regiões de interesse predefinidas. Assim, é possível automatizar a contagem (individualizada) de árvores nestas imagens, aumentando a eficiência da gestão em inventário florestal de eucalipto. Neste trabalho foram avaliados 20 classificadores para a tarefa de segmentação de imagens. Foi utilizada uma amostra real, de uma grande empresa na área de papel e celulose, para analisar o problema de contagem de árvores considerando um ambiente prático. Os resultados indicam que foi possível automatizar esta atividade com erro de 0.7\%, em particular, com estratégias baseadas em combinação de classificadores. Ainda, é apresentada uma visão geral do desempenho de cada classificador que pode ser utilizada como base para tomada de decisão em tarefas futuras.

\end{abstract}

\section{Introdução}

O cultivo de eucalipto representa um papel de destaque na economia brasileira \cite{soares2015competitividade}. O eucalipto tem uma característica econômica vantajosa relacionada à sua diversidade, pois pode ser empregado nas indústrias de papel, celulose, madeira industrializada e carvão vegetal. Além disso, tem um efeito importante no controle ambiental no que diz respeito ao seqüestro de $CO2$, uma vez que um hectare de eucalipto remove $60$ toneladas de $CO2$ da atmosfera \cite{eucaliptocarbono}. As empresas de setor têm reportados investimentos volumosos para viabilizar a \textit{silvicultura de precisão}, que é o processo para aumentar o grau de automatização e eficiência com operações envolvendo madeira florestal \cite{de2016qualidade}. 

No sentido de aumentar esta eficiência, nos últimos anos há um aumento considerável de uso de veículos aéreos não tripulados (VANTs) para apoiar monitoramento do inventário florestal \cite{chaves2015uso}. Dentre os principais objetivos, a detecção de falhas de plantio e contagem de árvores têm sido alvo de diversos estudos; uma vez que são importantes na estimativa de produção das empresas. Além disso, a visão aérea proporcionada pelos VANTs podem fornecer informações complementares às detectadas em solo, bem como coletar dados de grandes áreas em um curto espaço de tempo. Por propocionar baixo custo de coleta de dados quando comparado com alternativas tradicionais (\textit{e.g.} satélites e veículos tripulados), tal coleta é realizada com maior frequência o que gera uma quantidade maior de dados a serem analisados \cite{chaves2015uso}. Em particular, as imagens aéreas das florestas são um tipo de informação coletada, geralmente em alta resolução, que por serem georreferenciadas são combinadas em imagens maiores por meio de uma técnica denominada ``mosaicagem'' \cite{turner2012automated}, \textit{i.e.}, a fusão de imagens para recobrimento fotográfico de uma área de interesse \cite{laliberte2011multispectral}.

As imagens coletadas pelos VANTs são recursos importantes para o inventário florestal de eucalipto, em especial, na estimativa de quantidade de árvores. A técnica mais tradicional para esta estimativa é baseada indicadores de área basal, que são técnicas estatísticas para estimar a densidade florestal de uma área \cite{de2011modelos}. A popularidade desta técnica é baseada na ideia de que é inviável a contagem individual de árvores para o inventário florestal, sendo necessário realizar tal procedimento por técnicas de amostragem. Por outro lado, é reconhecido que a medição por área basal pode obter resultados insatisfatórios, geralmente devido à processos imprecisos de amostragem \cite{do1993amostragem}.

Com o aumento de recursos computacionais e avanços na área de aprendizado de máquina, bem como a alta disponibilidade das imagens áreas, diversas pesquisas têm apresentados resultados promissores sobre métodos para contagem individual de árvores \cite{brandtberg1998automated,katsch2006forest,gonccalves2016classification}. Nesse caso, a ideia principal é identificar padrões nas imagens que representam a copa de uma árvore (ou muda) de eucalipto (unidade arbórea) e contabilizar a ocorrência desses padrões na área de interesse. Métodos de \textbf{segmentação de imagens} são utilizados para identificar tais padrões com base em dois critérios \cite{ren2003learning}: (1) uniformidade intra-região, que ocorre quando elementos de uma mesma região da imagem têm alta similaridade de brilho, cor, forma e textura; e (2) contraste inter-região, que ocorre quando elementos em regiões diferentes têm baixa similaridade de brilho, cor, forma e textura. Os métodos de segmentação de imagens em geral são não supervisionados \cite{zhang2008image}, na qual heurísticas baseadas nos dois critérios acima são aplicadas para identificação de bordas ou com base em algoritmos de agrupamento de dados \cite{sonawane2015brief}. No entanto, para domínios específicos (como florestas de eucaliptos) é possível realizar segmentação supervisionada de imagens, com apoio de algoritmos de classificação de dados, em que uma amostra de dados é rotulada (como copa de árvores) para definir previamente o critério de uniformidade intra-região dos elementos de interesse \cite{sonka2014image}.

Em relação ao uso de classificadores para segmentação de imagens, a ideia é rotular uma amostra em duas classes, ``árvores'' e ``não árvores'', e então realizar a tarefa de segmentação. A saída é comumente uma imagem binarizada (\textit{e.g.} cor preta para classe ``árvores'' e cor branca para ``não árvores'') e assim um algoritmo de contagem é aplicado nessa imagem. Na área de aprendizado de máquina dezenas de algoritmos de classificação foram propostos, cada um com suas vantagens e desvantagens, sendo necessário um procedimento experimental para escolha do classificador \cite{shah2008performance}. 

Neste trabalho é apresentada uma avaliação de classificadores para segmentação de imagens, em particular, na tarefa de automatização de contagem de árvores em florestas de eucalipto. O trabalho é motivado por uma constatação de que não há um algoritmo de classificação que é eficaz em todos os tipos de imagens, sendo importante um procedimento de avaliação e seleção de classificadores. Até mesmo para um domínio específico (como imagens áreas das copas de eucalipto) há variações importantes na coleta das imagens, como luminosidade do dia (\textit{e.g.} ensolarado ou nublado), altura e angulação das fotos, características da região da plantação, bem como possíveis problemas na técnica de mosaicagem. As principais contribuições deste trabalho são destacadas a seguir:

\begin{itemize}
\item É apresentada uma avaliação empírica de 20 algoritmos de classificação para segmentação de imagens, em que o critério de avaliação é a taxa de erro na contagem de árvores de eucalipto. Embora tal resultado não possa ser generalizado para outras imagens, são discutidas as características dos algoritmos com melhor desempenho que podem ser úteis para apoiar a seleção de classificadores em aplicações futuras.
\item Também é analisado o uso de restrições de domínio na contagem de árvores, após o processo de segmentação, com base na distância esperada entre árvores para melhorar o processo de contagem. Esta técnica obtém melhores resultados do que as técnicas convencionais baseadas na contagem de segmentos com formatos predefinidos; que são elipses no caso de copas de eucaliptos.
\end{itemize}

O restante deste texto está organizado da seguinte maneira. Na seção II são apresentados os fundamentos deste trabalho, descrevendo pocesso de extração de características das imagens e algoritmos de classificação utilizados na tarefa de segmentação supervisionada. Na Seção III é apresentado o procedimento de avaliação de classificadores para segmentação de imagens com foco no inventário de florestas de eucalipto. Os resultados experimentais são discutidos na Seção IV. Por fim, limitações do trabalho e direção para trabalhos futuros são apresentados na Seção V.

\section{Fundamentos Básicos}

O processo de segmentação supervisionada de imagens depende inicialmente em representar a imagem em um formato adequado para a tarefa de classificação \cite{sonka2014image}. Tal representação contém as características extraídas das imagens. Assim, uma imagem é representada no modelo espaço-vetorial, em que cada possível segmento $s$ da imagem é um vetor $m$-dimensional $s = (f_1,f_2,...,f_m)$ e $f_i$ indica a relevância da característica $i$ no segmento $s$. Além disso, caso o segmento $s$ seja rotulado pelo usuário na classe ``árvore'' ou ``não árvore'', então fará parte do conjunto de treinamento para aprendizado do classificador.

Nas próximas seções é apresentado em mais detalhes o processo de extração de características, bem como uma breve descrição dos classificadores utilizados neste trabalho na tarefa de segmentação de imagens.

\subsection{Extração de Características das Imagens}

Uma imagem digital pode ser definida como uma matriz $M(x,y)$ de pixels com valores de intensidade, luminosidade e cor, associados a cada pixel. Para o processo de extração de características de uma imagem ou de um fragmento dela, é comum o uso de histogramas de cor, que é a representação numérica das características da imagem. A principal vantagem do uso de histogramas é a representação compacta das características da imagem \cite{sonka2014image}.

Um histograma geralmente é representado por um vetor contendo as variações de cores contidas na imagem, dadas em porcentagem, sendo invariante as transformações geométricas (escala, rotação e translação). As cores têm grande significância na indexação e recuperação de imagens. Podem ser representadas por diferentes padrões tais como RGB (red, green, blue) e HSI (hue, saturation, intensity); esse último mais próximo da percepção de cores da visão humana. Histogramas não possuem representatividade espacial dos pixels, sendo assim é possível que imagens diferentes possuam histogramas iguais ou próximos. Embora a cor possa ser facilmente utilizada como característica, há casos em que ela não pode sozinha representar característica, dado a isso os sistemas mais bem difundidos se utilizam de múltiplas características \cite{nixon2012feature}.

A textura é outro tipo de característica que pode ser suscintamente definida como a representação de padrões repetitivos em uma imagem de forma equânime. Podem ser analisadas por procedimento dentro de uma janela, denominada de análise estatística, ou se for feito no elemento da textura é denominado de analise estrutural. Analise estrutural se aplica a elementos claramente identificáveis, enquanto análise estatística a elementos mais pequenos e difíceis de identificar \cite{manjunath1996texture}. A análise da textura é baseada em formatos e regras descrevendo o posicionamento dos elementos relativos aos demais, como vizinhança e conexidade. Também são usadas regras sobre  densidade de elementos por unidade espacial, regularidade e homogeneidade. A segmentação baseada em textura determina as regiões que possuem textura uniforme, em que após identificação da textura um retângulo envolvente é usado para criar uma indexação do tipo R-Tree \cite{rosenfeld2014image}.

Além da cor e textura, também é possível caracterizar imagens com base nas formas dos seus elementos internos. É uma das tecnicas mais difíceis para extração de características devido à dificuldade em identificar tais formas, limitando a extração de características baseada em poucos objetos claramente identificados \cite{sonka2014image}. Em geral, abrange o pré-processamento da imagem para encontrar os objetos e detectar bordas podendo ser dificultado pela oclusão parcial de objetos e ruídos ou sombras. Há domínios que não apresentam formas pré-definidas, tais como tumores e mancha de pragas em folhas. Por outro lado temos domínios com formas geométricas definidas, por exemplo, localizar uma placa em um carro.

As características podem ser combinadas em um único vetor $m$-dimensional ou serem utilizadas separadamente para o treinamento do classificador. Trabalhos recentes na literatura indicam que uma combinação de características produz resultados mais satisfatórios do que o uso de apenas um tipo de característica \cite{nixon2012feature}.

\subsection{Métodos de Classificação}

A seguir são apresentados os métodos de classificação utilizados neste trabalho para apoiar a  segmentação de imagens visando a contagem de árvores de eucalipto. Para cada método é apresentada uma breve descrição com base na documentação apresentada na ferramenta Weka\cite{witten2016data} e FIJI-ImageJ \cite{schindelin2012fiji}, a qual contém referências para uma leitura mais detalhada desses algoritmos. 

\textit{$\bullet$ NaiveBayes:} fornece uma abordagem simples de aprendizado probabilístico, baseado no teorema de Bayes. Uma limitação do algoritmo é assumir que todas as características são independentes entre si durante a indução do classificador.

\textit{$\bullet$ Naive Bayes Updatable:} é uma versão incremental do NaiveBayes que é útil para grandes conjuntos de treinamento.

\textit{$\bullet$ Naive Bayes Multinonomial:} desenvolvido para dados com representações de alta dimensão, geralmente baseado em frequência de ocorrência das características, e com grande esparsidade (muitos elementos nulos ou zeros), como textos. O modelo multinomial captura informações de frequência de ocorrência, criando um vocabulário. Também permite realizar o aprendizado de forma incremental, sendo útil para grandes conjuntos de treinamento.

\textit{$\bullet$ SimpleLogistic:} constrói modelos de regressão logística linear. Para problemas de classificação, utiliza-se um limiar para definir a classe após o treinamento do regressor.

\textit{$\bullet$ Logistic:} constrói um modelo de regressão logística multinomial. Para problemas de classificação, utiliza-se um limiar para definir a classe após o treinamento do regressor.

\textit{$\bullet$ VotedPerceptron:} um classificador
baseado em uma lista de Perceptrons que, durante o treinamento, recebem pesos conforme a quantidade de objetos que conseguem classificador corretamente. Os pesos são utilizados, na etapa de teste, em uma estratégia de votação de cada Perceptron para a classificação final.

\textit{$\bullet$ MultilayerPerceptron:} MultilayerPerceptron é um tipo de Rede Neural que usa o algoritmo \textit{backpropagation} para classificar instâncias. Na implementação utilizada, os nós nessa rede usam a função sigmóide.

\textit{$\bullet$ RBFClassifier:} classificação usando redes de função de base radial, treinadas de forma totalmente supervisionada com minimização de erro quadrático. Na implementação utilizada, todos os atributos são normalizados na escala [0,1]. Os centros iniciais para as funções de base radial gaussianas são encontrados usando uma execução do algoritmo \textit{k-means}.

\textit{$\bullet$ SMO:} representa o algoritmo para Máquinas de Vetores de Suporte, baseado na teoria do aprendizado estatístico. Na implementação utilizada, todos os atributos são normalizados por padrão.  Também é utilizado um kernel polinomial, o que permite o aprendizado de classificadores não lineares.

\textit{$\bullet$ SMO LibLinear:} SMO LibLinear é uma biblioteca open source para classificação linear em larga escala. Utiliza como classificador base máquinas de vetores de suporte linear.
.

\textit{$\bullet$ FLDA:} constrói um função Discriminante Linear de Fisher. O limiar é selecionado de modo que o separador esteja a meio caminho entre os centróides das classes.
    
\textit{$\bullet$ IB1:} classificação com a técnica de vizinho mais próximo. Usa a distância euclidiana para encontrar a instância de treinamento mais próxima à instância de teste fornecida e prediz a mesma classe dessa instância de treinamento. Se várias instâncias tiverem a mesma (menor) distância para a instância de teste, a primeira encontrada será utilizada.
 
\textit{$\bullet$ RandomCommittee:} constroi um conjunto de classificadores base e a predição final é uma média linear das previsões geradas pelos classificadores base individuais. O classificador base comumente utilizado é o \textit{DecisionStump}.

\textit{$\bullet$ RandomSubSpace:} constrói um classificador que consiste em várias árvores construídas sistematicamente por seleção pseudo-aleatória de subconjuntos de características, isto é, árvores construídas em subespaços escolhidos aleatoriamente.

\textit{$\bullet$ RandomForest:} é um comitê de árvores de decisão (floresta), em que cada árvore é construída por meio de subconjunto de atributos e de instâncias. A classificação final é a moda da classificação de todas as árvores. Tal proposta tem a vantagem de evitar um modelo de classificação que faz um super ajuste nos dados, ou seja, melhora a tarefa de generalização.  

\textit{$\bullet$ FastRandomForest:} é uma versão otimizada da RandomForest no sentido de uso de memória. Dentre as otimizações, utiliza processamento paralelo via threads. Assim como a RandomForest, o resultado de classificação pode ser diferente em cada execução conforme a semente pseudo-aleatória. 

\textit{$\bullet$ PART:} contrói uma lista de regras de decisão com base em uma árvore de decisão, na qual em cada iteração a ``melhor folha'' atual é transformada em uma regra.

\textit{$\bullet$ DecisionStump:} é uma árvore de decisão com apenas um nível. Em geral, cada atributo gera uma regra caso sua entropia seja baixa, ou seja, caso o atributo seja discriminativo para as classes. Por ser simples e rápido, os algoritmos baseados em comitês usualmente utilizam por padrão este classificador como base.
   
\textit{$\bullet$ J48:} é uma implementação baseada no algoritmo C4.5 para construção de árvore de decisão, com base em ganho de informação. A árvore de decisão pode ser podada ou não.

\textit{$\bullet$ LMT}: classificador para a construção de ``árvores de modelo logístico'', que são árvores de classificação com funções de regressão logística nas folhas. 

\section{Avaliação de Classificadores para Segmentação de Imagens}
\begin{figure*}[htbp!]
    \centering
    \fbox{\includegraphics[width=1\textwidth]{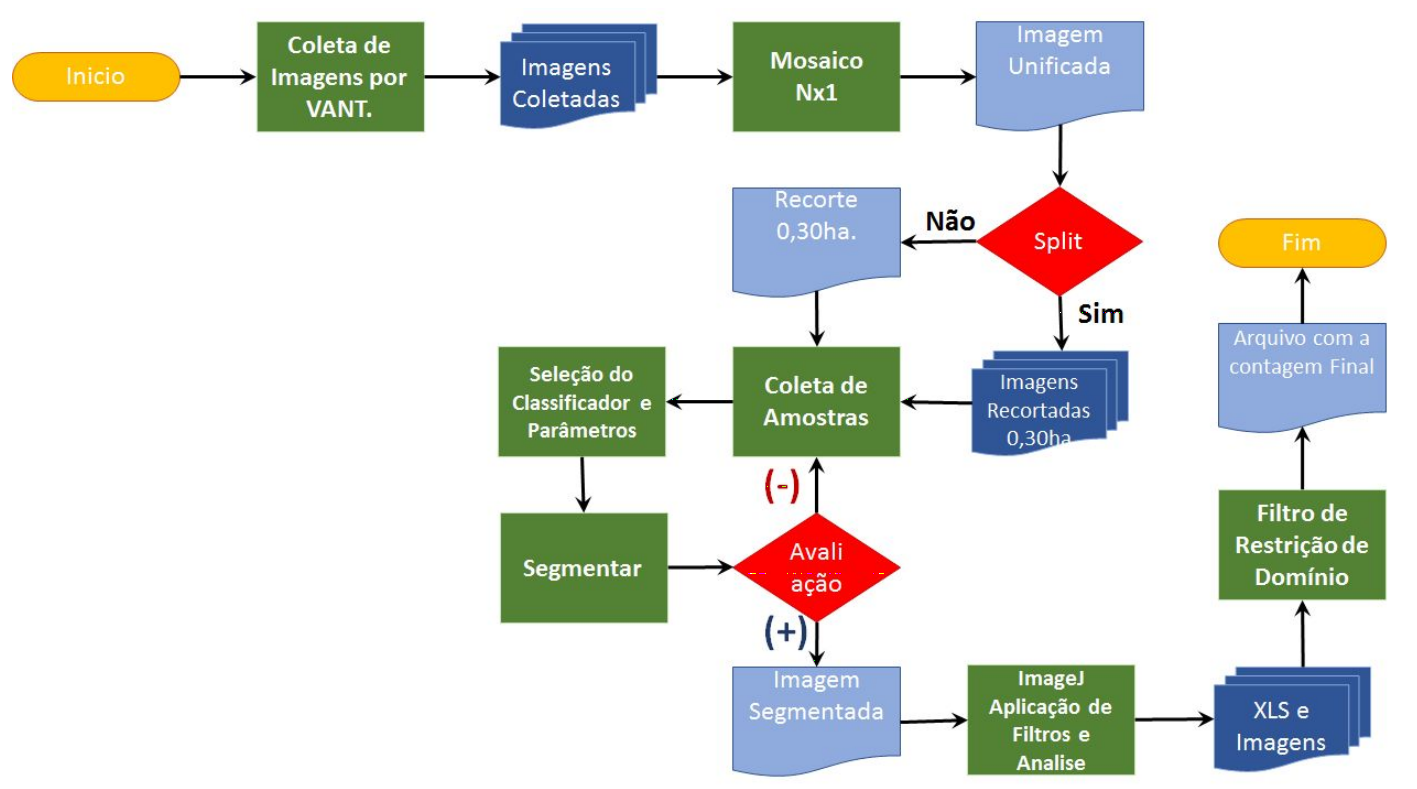}}
    \caption{Passos da metodologia utilizada para o processo de inventário florestal de eucalipto, em que a etapa de contagem de árvores é apoiada por técnicas de segmentação de imagem via métodos de classificação.}
	\label{metodologia}
\end{figure*}
Conforme comentado anteriormente, aplicações para apoiar o inventário florestal têm recebido grande atenção em empresas do ramo de papel e celulose. Neste processo, aumentar a precisão na contagem de árvores de eucalipto é fundamental, pois permite melhorar a estimativa da produção futura. O desenvolvimento deste trabalho visa apoiar a metodologia adotada em uma das principais empresas brasileiras no setor, localizada na região de Três Lagoas - MS. Na Figura \ref{metodologia} são ilustrados os passos da metodologia utilizada para apoiar o processo de inventário florestal de eucalipto.

\begin{figure*}[htbp!]
  \centering
    \fbox{\includegraphics[width=1\textwidth]{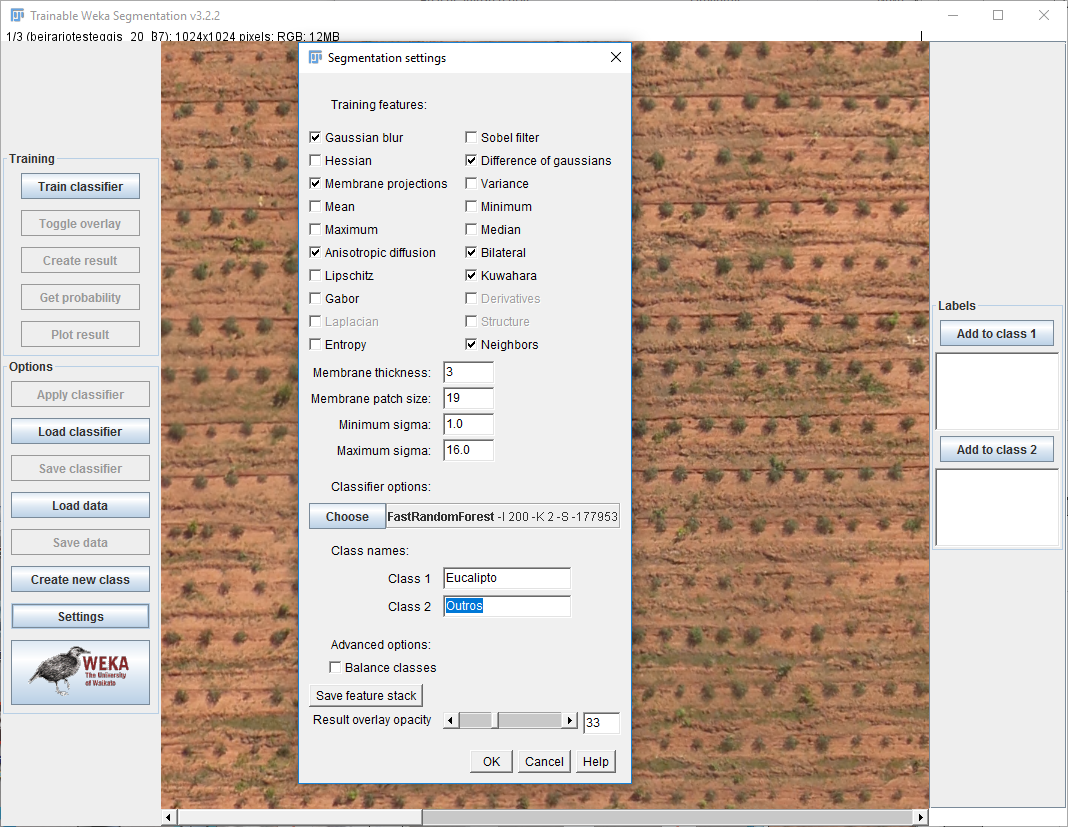}}
      \caption{Interface do FIJI-ImageJ integrado ao Weka. O usuário seleciona um conjunto de técnicas para extração de características da imagem. Em seguida, seleciona regiões da imagem que farão parte amostra anotada (classes). Por fim, o algoritmo de classificação e seus parâmetros são selecionados para iniciar a etapa de segmentação supervisionada da imagem.}
      	\label{etapa1}
\end{figure*}

\begin{figure}[htbp!]
  \centering
    \fbox{\includegraphics[width=0.45\textwidth]{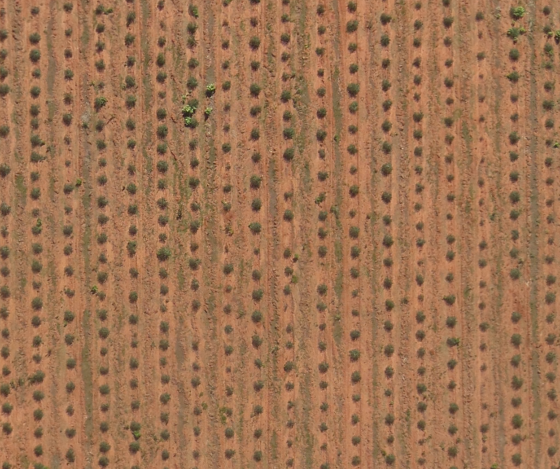}}
      \caption{Imagem área com plantação de mudas de eucalipto.}
      	\label{original}
\end{figure}

\begin{figure}[htbp!]
  \centering
    \fbox{\includegraphics[width=0.45\textwidth]{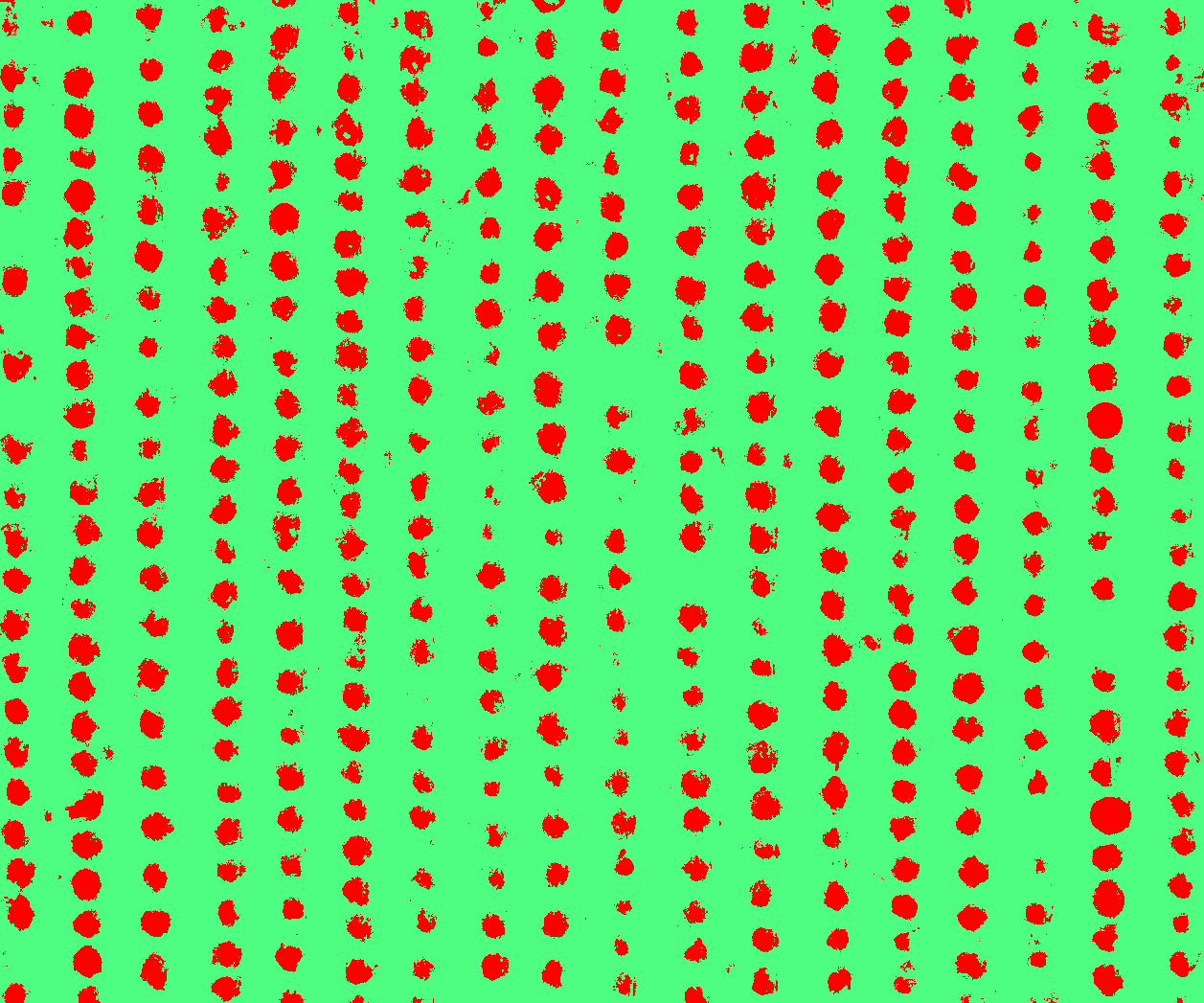}}
      \caption{Imagem resultante da segmentação supervisionada.}
      	\label{etapa2}
\end{figure}

\begin{figure}[htbp!]
  \centering
  \fbox{\includegraphics[width=0.45\textwidth]{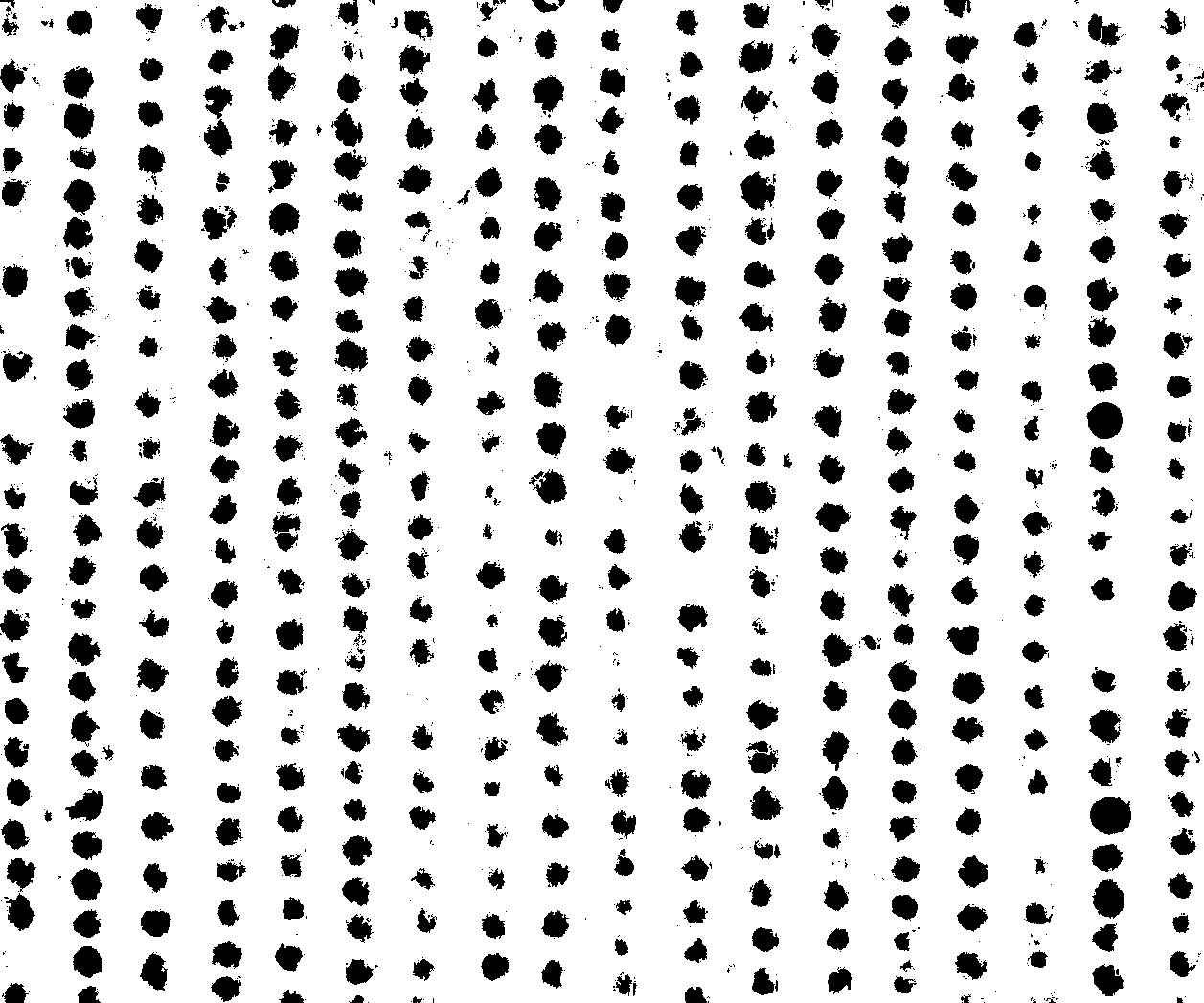}}
  \caption{Binazariação da imagem segmentada e aplicação de filtros de realce de elipses para apoiar a contagem de árvores.}
    \label{etapa3}
\end{figure}

\begin{figure}[htbp!]
  \centering
    \fbox{\includegraphics[width=0.45\textwidth]{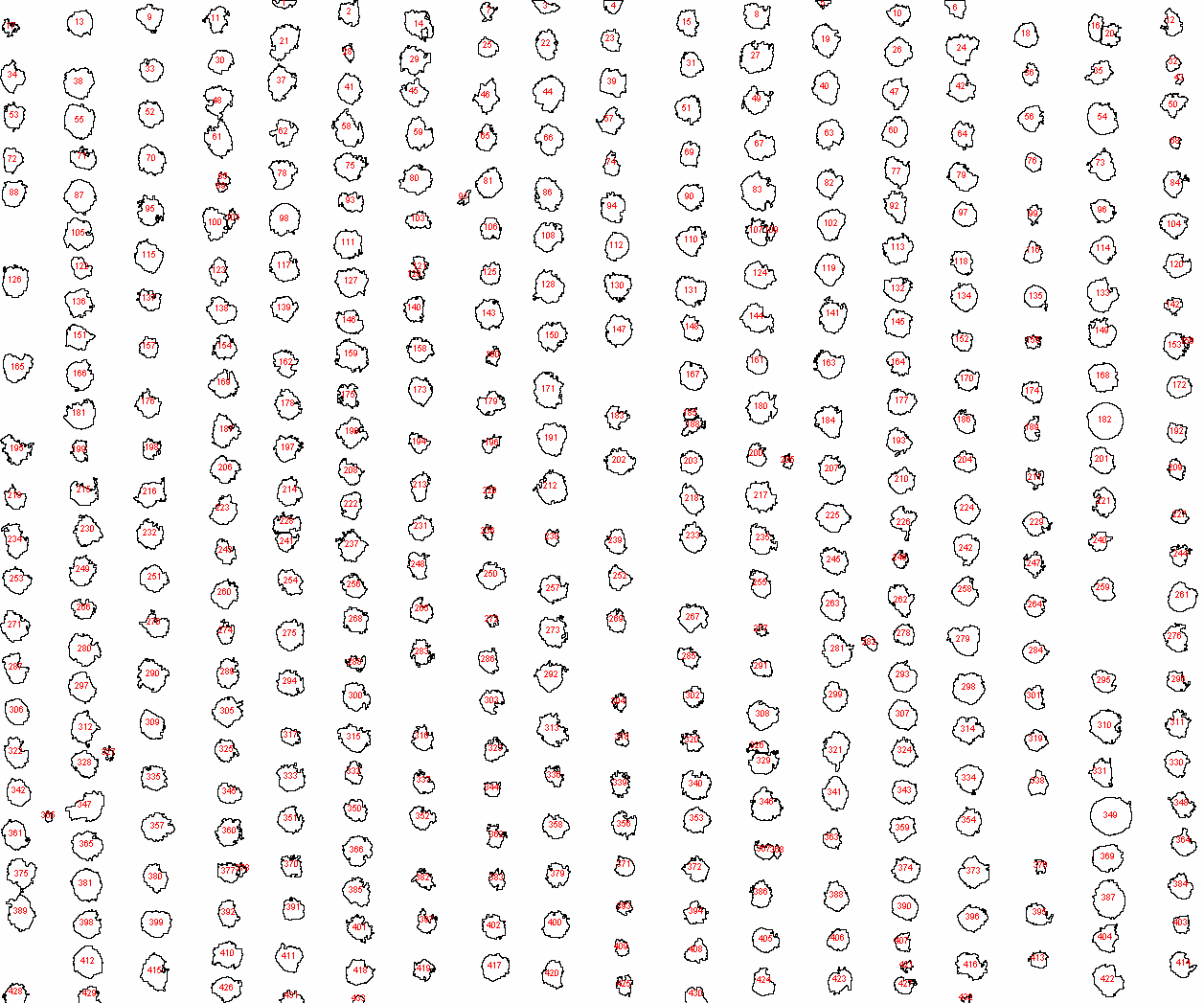}}
      \caption{Imagem resultante do processo de contagem de árvores. Essas informações também são apresentadas em dados tabulados para serem importados pelos sistemas de informações geográficas.}
  \label{etapa4}
\end{figure}

O processo é iniciado por meio da coleta de imagens por VANTs. O processo de fusão de imagens é aplicado ao final da coleta, gerando um mosaico (imagem unificada) da área de interesse. Em seguida, os técnicos da área de inventário florestal realizam recortes na imagem para (i) remover plantações que estão na borda da área de interesse e (ii) definir uma amostra para o processo de segmentação de imagens. Este trabalho apoia diretamente a etapa de seleção do classificador apropriado para a segmentação, considerando a imagem coletada. A imagem segmentada é utilizada como entrada para algoritmos que realizam o processamento da imagem e contagem de árvores, disponíveis na ferramenta FIJI-ImageJ. A ideia desses algoritmos é identificar segmentos similares à elipses, que possuem maiores chances de serem copas (ou mudas) de eucalipto. Essas informações são tabuladas em uma planilha indicando a posição na imagem em que cada unidade arbórea foi contabilizada. Por fim, é proposto neste trabalho o uso de um filtro de restrições de domínio, que remove contagens inválidas. Nesse caso, informações predefinidas sobre a distância mínima entre árvores (ou mudas) são utilizadas como restrições. Enfim, a eficácia desta metodologia depende da integração de vários softwares utilizados no processo:

\begin{itemize}
\item QGIS: sistema de informação geográfica que permite analisar imagens georreferenciadas coletadas pelos VANTs;
\item FIJI-ImageJ: ambiente para processamento de imagens que auxilia tanto na etapa de extração de características das imagens quanto na contagem de árvores após a segmentação; e
\item Weka: ambiente de aprendizado de máquina com dezenas de algoritmos de classificação e utilitários para avaliação de classificadores.
\end{itemize}

Para melhor compreensão desta metodologia, a seguir são ilustrados os passos principais. Na Figura \ref{etapa1}, é ilustrada a etapa em que o usuário deve identificar quais regiões das imagens representam as classes de interesse. Na prática, o usuário seleciona na própria imagem as regiões que representam mudas (ou copas) de eucalipto para formar a classe de interesse e outras regiões da imagem para representar a classe ``outros (ou não árvore)''. O usuário também deve identificar quais as técnicas de processamento de imagem serão utilizadas para extração das características, geralmente baseada em cores e texturas. Por fim, o classificador para segmentação supervisionada, bem como seus parâmetros devem ser definidos. A ferramenta apresenta os parâmetros melhor avaliados na literatura como padrão.

Considere uma imagem com mudas de eucalipto conforme ilustrado na Figura \ref{original}. Na Figura \ref{etapa2} é ilustrada esta imagem segmentada por meio de um classificador. Note que a classe de interesse é apresentada por elipses, que visa futuramente facilitar o processo de contagem de árvores. Para finalizar esse processo, ainda é necessário transformar a imagem para um formato binário. Em seguida, aplicam-se os filtros denominados ``\textit{Fill Holes}'' e ``\textit{Watershed}''. O primeiro tem o objetivo de realçar formas de elipses, preenchendo seu conteúdo com pixels preto. O segundo permite realçar as bordas de divisão entre as elipses identificadas. A imagem resultante é ilustrada na Figura \ref{etapa3}.

Para finalizar o processo aplica-se um algoritmo para contagem de árvores propriamente dito, que é baseado na análise de particulas. A ideia é buscar agrupamentos de pixels, em formato de elipses, definindo como parâmetro um raio mínimo (nesse trabalho o minino é 50px). Assim, obtemos uma imagem com a eliminação dos grupos de pixels fora do limite especificado para o domínio, como ilustrado na Figura \ref{etapa4}. As informações relacionadas à contagem das árvores são apresentadas de forma tabulada, contendo um identificador e as coordenadas do centro de cada agrupamento de pixels identificados como a classe de interesse.

Por fim, neste trabalho, é proposto o uso de regras com restrições de domínio (como a distância mínima conhecida entre árvores) podem ser empregadas nesses dados tabulados, a fim de remover possíveis contagens inválidas. Nesse caso, o usuário deve informar restrições do domínio previamente. O algoritmo de filtro verifica, para cada árvore, se (i) o raio da copa (ou muda) satisfaz às restrições; se (ii) existem árvores vizinhas a uma distância menor do que a permitida; e se (iii) existe três árvores consecutivas formam possuem uma tendência linear devido ao estilo de plantação. Tais restrições permitem identificar ruídos que foram incorretamente consideradas como árvores. É importante ressaltar que a saída desse processo é uma planilha tabulada que pode ser importada novamente para o sistema de informação geográfica e, assim, realçar aos usuários as árvores contabilizadas na área de interesse.

\section{Avaliação Experimental}

A avaliação experimental consiste em analisar o erro obtido em relação à contagem de árvores no final do processo, variando-se o classificador utilizado para segmentação da imagem.

Um problema crítico para este tipo de avaliação é a obtenção de uma imagem já contabilizada e validada por humanos para ser utilizada como referência (conjunto verdade). Conforme comentado anteriormente, o processo manual de contagem de árvores é inviável. Uma alternativa comumente utilizada por outros trabalhos é comparar a contabilização obtida via segmentação de imagens com a contabilização obtida via amostragem por área basal. No entanto, acreditamos que essa comparação poderia levar a erros, uma vez que o conjunto de referência seria obtido por uma estratégia não confiável. Desse modo, neste trabalho optou-se por contabilizar (de fato) manualmente as árvores de uma região de interesse para construir um conjunto de referência. Embora a amostra para a avaliação experimental seja reduzida, há a vantagem de que a análise dos resultados é mais próxima de uma aplicação real.

\begin{table*}[htbp]
\centering
\caption{Visão geral da avaliação dos classificadores para segmentação de imagem. O erro de contagem é a diferença entre o número de mudas e o número de árvores identificadas com base na imagem segmentada, sem a aplicação de filtro com restrição de domínio. Os tempos são dados em segundos.}
\begin{tabular}{l|c|c|c|c|c}
\hline
\hline
\textbf{Classificador} & \multicolumn{1}{l|}{\textbf{Tempo de Treino}} & \multicolumn{1}{l|}{\textbf{Tempo da Segmentação}} & \multicolumn{1}{l|}{\textbf{Tempo Total}} & \multicolumn{1}{l|}{\textbf{Erro}} & \multicolumn{1}{l}{\textbf{Erro(\%)}} \\ \hline
\hline
{Decision Stump} & {0.2} & {0.2} & {0.4} & {-12} & {2.9\%} \\ \hline
{RBFClassifier} & {14.7} & {1.3} & {16.0} & {+15} & {3.6\%} \\ \hline
{FastRandomForest} & {5.6} & {39.9} & {45.4} & {+21} & {5.1\%} \\ \hline
{RandomCommittee} & {42.3} & {34.0} & {76.3} & {+57} & {13.8\%} \\ \hline
{LibLINEAR} & {12.9} & {1.3} & {14.2} & {+63} & {15.3\%} \\ \hline
{RandomForest} & {13.8} & {10.2} & {24.0} & {+66} & {16.0\%} \\ \hline
{RandomSubSpace} & {5.8} & {1.9} & {7.8} & {+107} & {26.0\%} \\ \hline
{Naive Bayes} & {0.3} & {10.4} & {10.7} & {+135} & {32.8\%} \\ \hline
{Naive Bayes Updatable} & {0.2} & {10.2} & {10.4} & {+135} & {32.8\%} \\ \hline
{SMO} & {55.5} & {1.1} & {56.6} & {143} & {34.7\%} \\ \hline
{MultilayerPerceptron} & {498.6} & {6.0} & {504.6} & {+151} & {36.7\%} \\ \hline
{FLDA} & {0.2} & {1.1} & {1.3} & {+156} & {37.9\%} \\ \hline
{IB1} & {0.0} & {7033.0} & {7033.0} & {+166} & {40.3\%} \\ \hline
{PART} & {9.8} & {0.6} & {10.4} & {+170} & {41.3\%} \\ \hline
{SimpleLogistic} & {31.8} & {1.2} & {33.0} & {+177} & {43.0\%} \\ \hline
{LMT} & {188.8} & {0.8} & {189.6} & {+178} & {43.2\%} \\ \hline
{VotedPerceptron} & {13.4} & {604.1} & {617.5} & {+188} & {45.6\%} \\ \hline
{Logistic} & {3.0} & {1.1} & {4.1} & {+200} & {48.5\%} \\ \hline
{J48} & {3.8} & {0.3} & {4.1} & {+294} & {71.4\%} \\ \hline
{Naive Bayes Multinomial} & {0.0} & {1.8} & {1.9} & {+430} & {104.4\%} \\ \hline
\hline
\end{tabular}
\label{resultados}
\end{table*}

Todos os classificadores descritos na Seção II foram utilizados nesta avaliação. Adotou-se os parâmetros definidos como padrão de cada classificador, que usualmente obtém bons resultados conforme descrito na documentação das ferramentas (FIJI-ImageJ integrado ao Weka). 

A imagem de referência contém 412 mudas de eucalipto. Foram analisados dois aspectos durante a avaliação: taxa de erro na contagem e tempo em segundos (treinamento e teste) da segmentação. O computador utilizado para os experimentos possui quatro núcleos de 3Ghz e 8GB de memória RAM.

Na Tabela \ref{resultados} é apresentado um resumo geral da avaliação experimental. Os tempos de treinamento, classificação, e total são fornecidos em segundos. O erro de contagem é a diferença entre o número real de mudas e o número de árvores identificadas com base na imagem segmentada, sem a aplicação de filtro com restrição de domínio. Quando é contabilizado um número maior de árvores do que o real, então o erro é precedido pelo sinal de $+$. Caso contrário, é precedido pelo sinal de $-$. Cada classificador foi ordenado conforme o erro de contagem, do menor para o maior.

Uma análise dos principais pontos dos resultados experimentais é listada abaixo:

\begin{itemize}
\item Embora o classificador DecisionStump tenha obtido o menor erro, apresenta um efeito indesejado que é subestimar o número de árvores. Isso significa que o filtro de restrições de domínio para eliminação de ruídos contabilizado como árvore não terá o efeito desejado quando for aplicado.
\item O classificador RBFClassifier obeve resultados promissores. Uma possível explicação é relacionada à sua estratégia de inicialização de parâmetros, baseado em uma execução do algoritmo \textit{k-means}. No contexto da segmentação, o \textit{k-means} permite identificar grupos com regiões uniformes, ou seja, é um processo de segmentação não supervisionada. A rede neural RBF, então, durante seu treinamento basicamente realiza um ajuste dessa segmentação utilizando a minimização do erro quadrático nas classes fornecidas.
\item Estratégias baseados na combinação de vários classificadores obtiveram resultados satisfatórios, como FastRandomForest, RandomForest, RandomCommittee e RandomSubSpace. Tal resultado é baseado em dois fatores: (i) aumento da capacidade de generalização por utilizar subconjuntos de instâncias e atributos e (ii) maior chance em explorar de forma independente cada tipo de característica extraída das imagens (cores e texturas) ao selecionar subconjuntos de características.
\item Estratégias baseadas em aprendizado estatístico (SMO) e redes MLP apresentaram um erro significativo. No entanto, existe uma grande combinação de parâmetros (e topologias no caso de MLP) que poderiam ser testadas para minimizar o erro. Do ponto de vista dos autores deste trabalho, estratégias livres de parâmetros ou que determinam automaticamente os parâmetros são potencialmente mais úteis em cenários com grande diversidade de imagens.
\item Estratégias baseadas em regressão logística não apresentaram resultados satisfatórios para segmentação de imagens, principalmente devido à alta dimensionalidade deste tipo de problema.
\item Estratégias baseadas em vizinhos mais próximos (IB1) são ineficientes no quesito de tempo e também não apresentaram resultados satisfatórios. As imagens em alta resolução e a dificuldade em definir uma medida de distância apropriada inviabiliza seu uso prático.
\item Vale destacar os resultados negativos do classificador Naive Bayes Multinomial. Esse resultado é esperado uma vez que as características para representação do problema não são esparsas, fazendo com que o modelo multinomial não compute corretamente as probabilidades das classes.
\end{itemize}

Após esta etapa da avaliação, foram selecionados os três melhores classificadores, conforme o erro de contagem, para aplicação do filtro com restrições de domínio. Nesse caso, o erro do classificador \textit{DecisionStump} aumentou para $5.6$\%, após a remoção de $11$ árvores incorretamente contabilizadas. Já o erro do classificador \textit{RBFClassifier} foi reduzido para $1.9$\% após remoção de $23$ árvores incorretamente contabilizadas. Por fim, o erro do classificador \textit{FastRandomForest} foi reduzido para $0.7$\% após remoção de $18$ árvores incorretamente contabilizadas. Assim, considerando os resultados experimentais, o classificador \textit{FastRandomForest} é uma opção promissora, tanto em relação à capacidade de segmentação (principalmente na presença de diferentes tipos de característica de imagens) quanto em relação ao custo computacional.

Por fim, no Apêndice A são apresentadas imagens segmentadas obtidas por cada classificador para possibilitar uma análise visual dos resultados.

\section{Considerações Finais}

A avaliação experimental para escolha de classificadores na segmentação de imagens é uma etapa importante para a contagem de árvores, uma vez que não há um método de classificação que será promissor em todas as variações de imagem. Nesse sentido, destaca-se o uso de métodos baseados em combinação de classificadores, que se mostraram competitivos quando comparados com classificadores usualmente utilizados na área, como Redes MLP e Máquinas de Vetores de Suporte.

Uma limitação deste trabalho é o uso de uma amostra reduzida para treinamento dos modelos, dada a reconhecida dificuldade em contabilizar manualmente grandes áreas de florestas de eucalipto. No entanto, a imagem de referência escolhida é representativa para a área, e será disponibilizada para outros trabalhos.

Como direção para trabalho futuro, espera-se incluir classificadores baseado em aprendizado profundo (\textit{Deep Learning}) na tarefa de contagem de árvores, uma vez que tem obtidos resultados interessantes em aprendizado de máquina envolvendo imagens.

\section{Agradecimentos}

Os autores agradecem à valiosa contribuição de Luiz Henrique Terezan (Supervisor de Qualidade Florestal na Eldorado Brasil Celulose S/A) para o desenvolvimento deste trabalho.

\bibliography{referencias}{}

\begin{thebibliography}{10}

\bibitem{soares2015competitividade}
N.~S. Soares, M.~L.~d. Silva, and M.~d.~M. Pires, ``Competitividade da
  produ{\c{c}}{\~a}o de eucalipto no brasil,'' {\em Revista de Pol{\'\i}tica
  Agr{\'\i}cola}, vol.~24, no.~1, pp.~112--124, 2015.

\bibitem{eucaliptocarbono}
``{Confederação da Agricultura e Pecuária do Brasil. Plantio de eucalipto no
  Brasil}. brasília,'' 2011.

\bibitem{de2016qualidade}
L.~F. Castro~Galizia, G.~A. Ramiro, C.~J. de~Carvalho~Rosa, C.~D.
  Operacional-Fibria, and A.~D. Operacional-Fibria, ``Qualidade das atividades
  silviculturais e silvicultura de precis{\~a}o,'' {\em S{\'e}rie T{\'e}cnica
  IPEF}, vol.~24, no.~45, 2016.

\bibitem{chaves2015uso}
A.~A. Chaves and R.~A. La~Scalea, ``Uso de vants e processamento digital de
  imagens para a quantifica{\c{c}}{\~a}o de {\'a}reas de solo e de
  vegeta{\c{c}}{\~a}o,'' {\em Anais XVII Simp{\'o}sio Brasileiro de
  Sensoriamento Remoto-SBSR, Jo{\~a}o Pessoa-PB, Brasil}, vol.~25, 2015.

\bibitem{turner2012automated}
D.~Turner, A.~Lucieer, and C.~Watson, ``An automated technique for generating
  georectified mosaics from ultra-high resolution unmanned aerial vehicle (uav)
  imagery, based on structure from motion (sfm) point clouds,'' {\em Remote
  Sensing}, vol.~4, no.~5, pp.~1392--1410, 2012.

\bibitem{laliberte2011multispectral}
A.~S. Laliberte, M.~A. Goforth, C.~M. Steele, and A.~Rango, ``Multispectral
  remote sensing from unmanned aircraft: Image processing workflows and
  applications for rangeland environments,'' {\em Remote Sensing}, vol.~3,
  no.~11, pp.~2529--2551, 2011.

\bibitem{de2011modelos}
S.~P{\'a}dua~Chaves, N.~C. Carvalho, F.~Fonseca, L.~A. C.~B. Silva, A.~R.
  de~Mendon{\c{c}}a, and M.~P. de~Lima, ``Modelos n{\~a}o lineares
  generalizados aplicados na predi{\c{c}}{\~a}o da {\'a}rea basal e volume de
  eucalyptus clonal,'' {\em Cerne, Lavras}, vol.~17, no.~4, pp.~541--548, 2011.

\bibitem{do1993amostragem}
H.~T.~Z. do~Couto, N.~L.~M. Bastos, and J.~S. de~Lacerda, ``A amostragem por
  pontos na estimativa de {\'a}rea basal em povoamentos de eucalyptus,'' {\em
  IPEF}, no.~46, pp.~86--95, 1993.

\bibitem{brandtberg1998automated}
T.~Brandtberg and F.~Walter, ``Automated delineation of individual tree crowns
  in high spatial resolution aerial images by multiple-scale analysis,'' {\em
  Machine Vision and Applications}, vol.~11, no.~2, pp.~64--73, 1998.

\bibitem{katsch2006forest}
C.~K{\"a}tsch and A.~Kunneke, ``Forest inventory in the digital remote sensing
  age,'' {\em Southern African Forestry Journal}, vol.~206, no.~1, pp.~43--49,
  2006.

\bibitem{gonccalves2016classification}
W.~G. Gon{\c{c}}alves, H.~M.~C. Ribeiro, J.~A. S.~d. S{\'a}, G.~P. Morales,
  H.~R. Ferreira~Filho, and A.~d.~C. Almeida, ``Classification of forest types
  using artificial neural networks and remote sensing data,'' {\em Revista
  Ambiente \& {\'A}gua}, vol.~11, no.~3, pp.~612--624, 2016.

\bibitem{ren2003learning}
X.~Ren and J.~Malik, ``Learning a classification model for segmentation.,'' in
  {\em ICCV}, vol.~1, pp.~10--17, 2003.

\bibitem{zhang2008image}
H.~Zhang, J.~E. Fritts, and S.~A. Goldman, ``Image segmentation evaluation: A
  survey of unsupervised methods,'' {\em Computer vision and image
  understanding}, vol.~110, no.~2, pp.~260--280, 2008.

\bibitem{sonawane2015brief}
M.~Sonawane and C.~Dhawale, ``A brief survey on image segmentation methods,''
  in {\em IJCA Proceedings on National conference on Digital Image and Signal
  Processing}, Citeseer, 2015.

\bibitem{sonka2014image}
M.~Sonka, V.~Hlavac, and R.~Boyle, {\em Image processing, analysis, and machine
  vision}.
\newblock Cengage Learning, 2014.

\bibitem{shah2008performance}
S.~K. Shah, ``Performance modeling and algorithm characterization for robust
  image segmentation,'' {\em International Journal of Computer Vision},
  vol.~80, no.~1, pp.~92--103, 2008.

\bibitem{nixon2012feature}
M.~S. Nixon and A.~S. Aguado, {\em Feature extraction \& image processing for
  computer vision}.
\newblock Academic Press, 2012.

\bibitem{manjunath1996texture}
B.~S. Manjunath and W.-Y. Ma, ``Texture features for browsing and retrieval of
  image data,'' {\em IEEE Transactions on pattern analysis and machine
  intelligence}, vol.~18, no.~8, pp.~837--842, 1996.

\bibitem{rosenfeld2014image}
A.~Rosenfeld, {\em Image modeling}.
\newblock Academic Press, 2014.

\bibitem{witten2016data}
I.~H. Witten, E.~Frank, M.~A. Hall, and C.~J. Pal, {\em Data Mining: Practical
  machine learning tools and techniques}.
\newblock Morgan Kaufmann, 2016.

\bibitem{schindelin2012fiji}
J.~Schindelin, I.~Arganda-Carreras, E.~Frise, V.~Kaynig, M.~Longair,
  T.~Pietzsch, S.~Preibisch, C.~Rueden, S.~Saalfeld, B.~Schmid, {\em et~al.},
  ``Fiji: an open-source platform for biological-image analysis,'' {\em Nature
  methods}, vol.~9, no.~7, pp.~676--682, 2012.

\end{thebibliography}
\bibliographystyle{ieeetr}

\begin{table*}
\centering
\begin{tabular}{c}
\large Apêndice A \\
\end{tabular}
\end{table*}

\begin{figure*}[htbp!]
  \centering
    \includegraphics[width=1\textwidth,page=1]{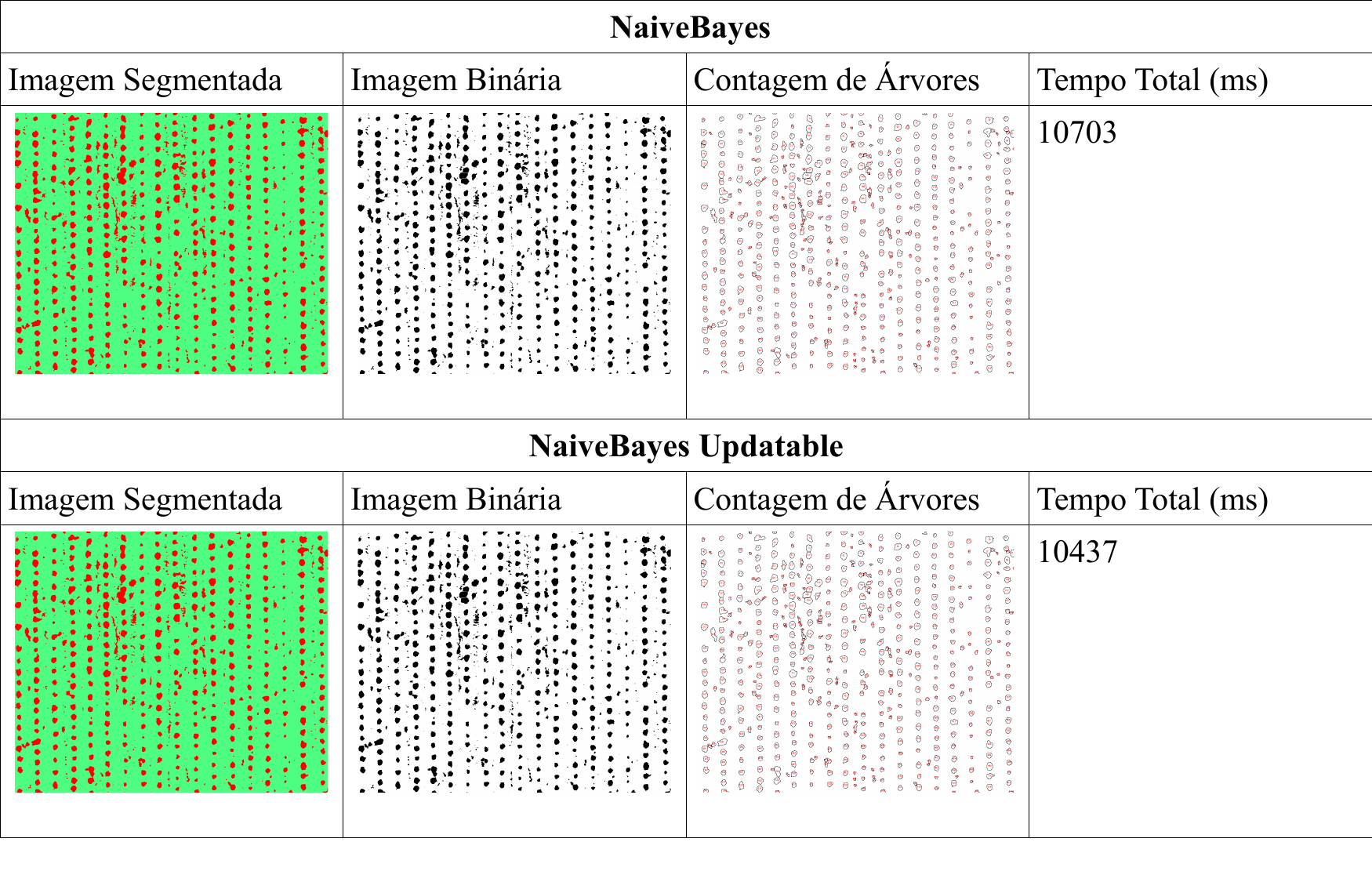}
\end{figure*}

\begin{figure*}[htbp!]
  \centering
    \includegraphics[width=1\textwidth,page=2]{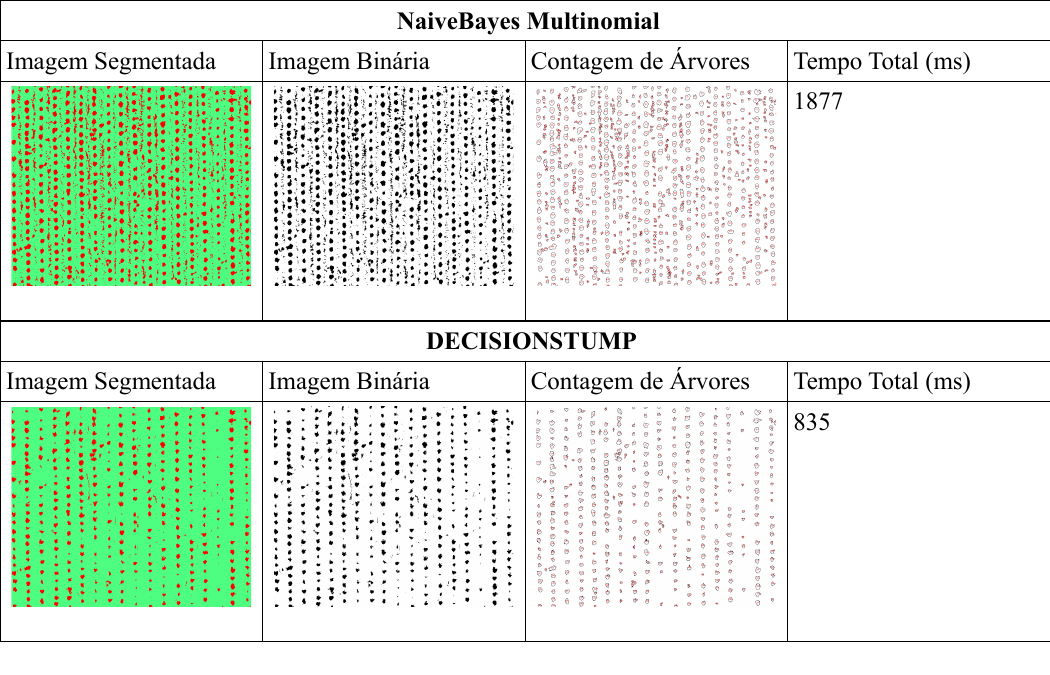}
\end{figure*}

\begin{figure*}[htbp!]
  \centering
    \includegraphics[width=1\textwidth,page=3]{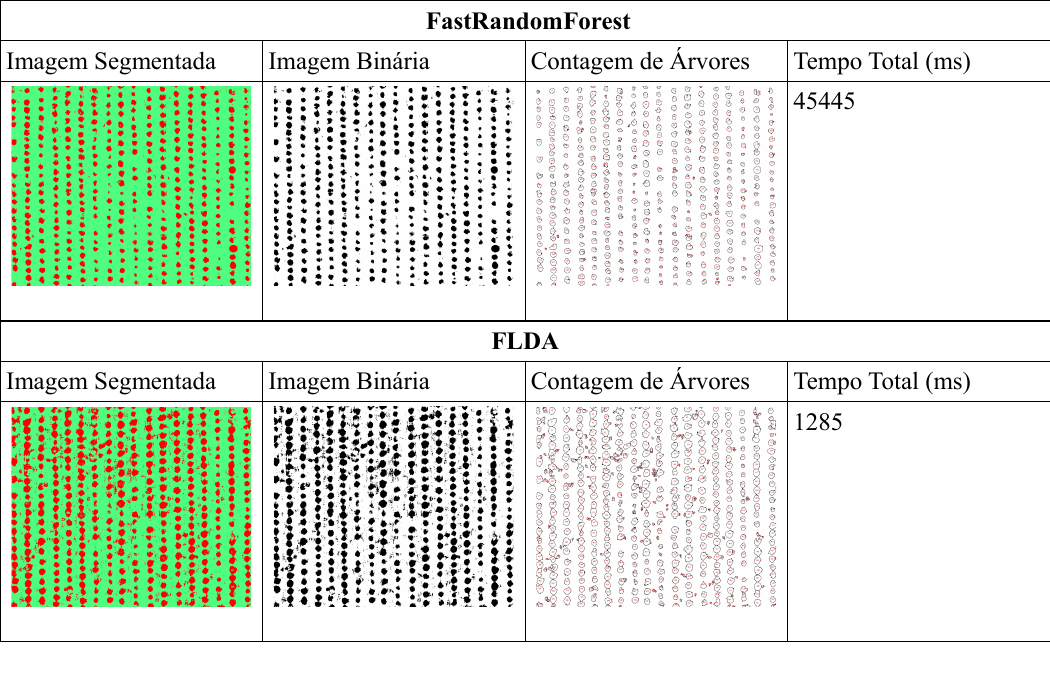}
\end{figure*}

\begin{figure*}[htbp!]
  \centering
    \includegraphics[width=1\textwidth,page=4]{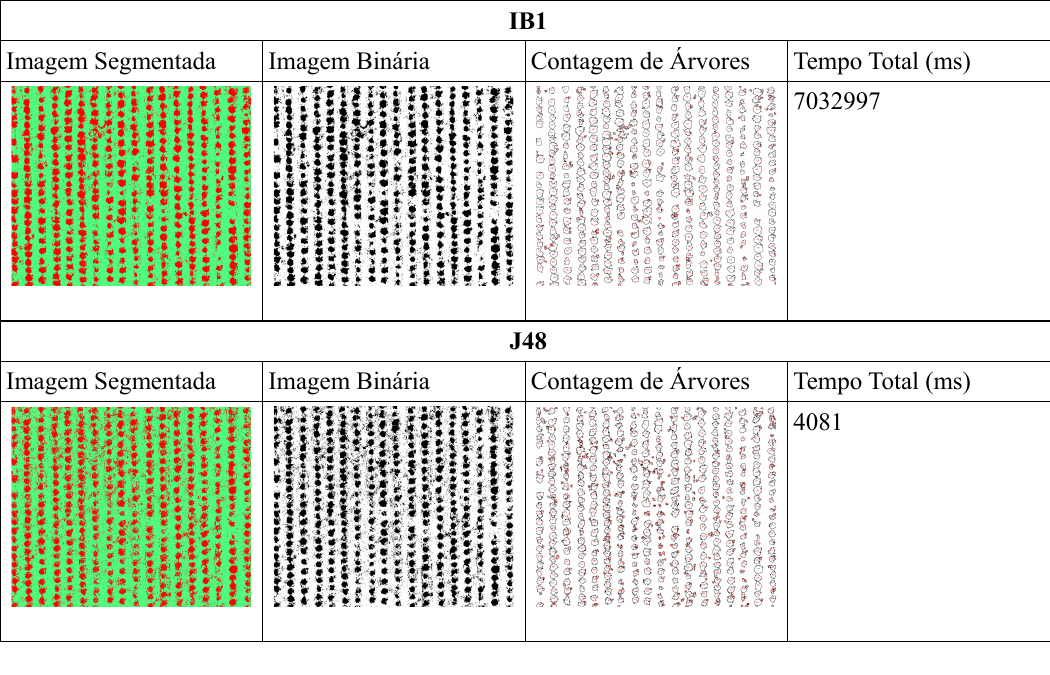}
\end{figure*}

\begin{figure*}[htbp!]
  \centering
    \includegraphics[width=1\textwidth,page=5]{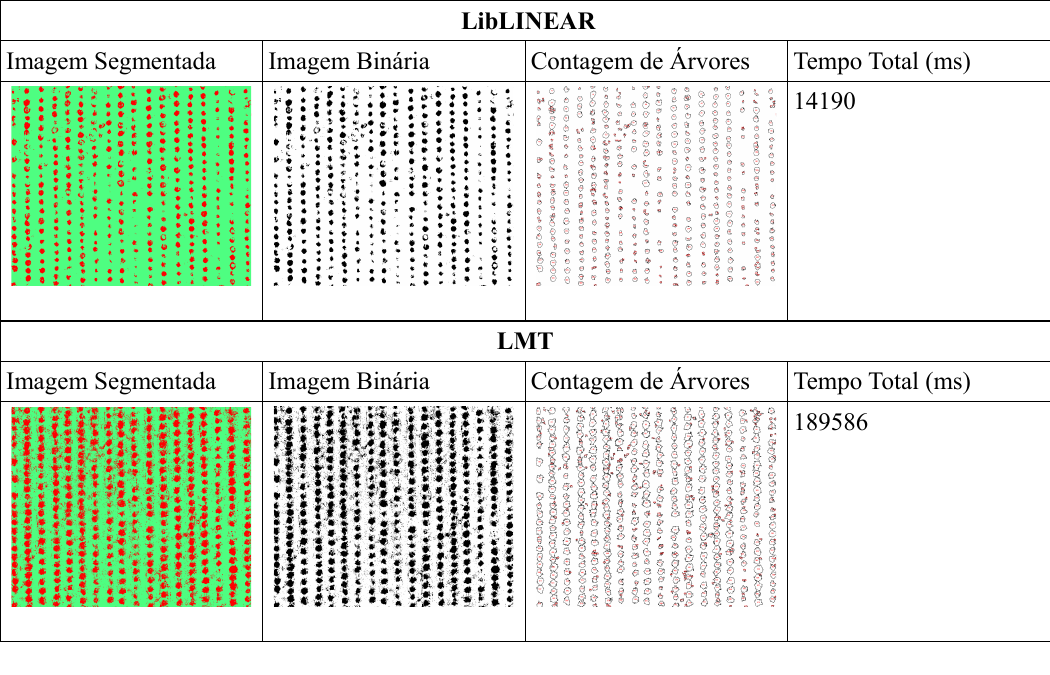}
\end{figure*}

\begin{figure*}[htbp!]
  \centering
    \includegraphics[width=1\textwidth,page=6]{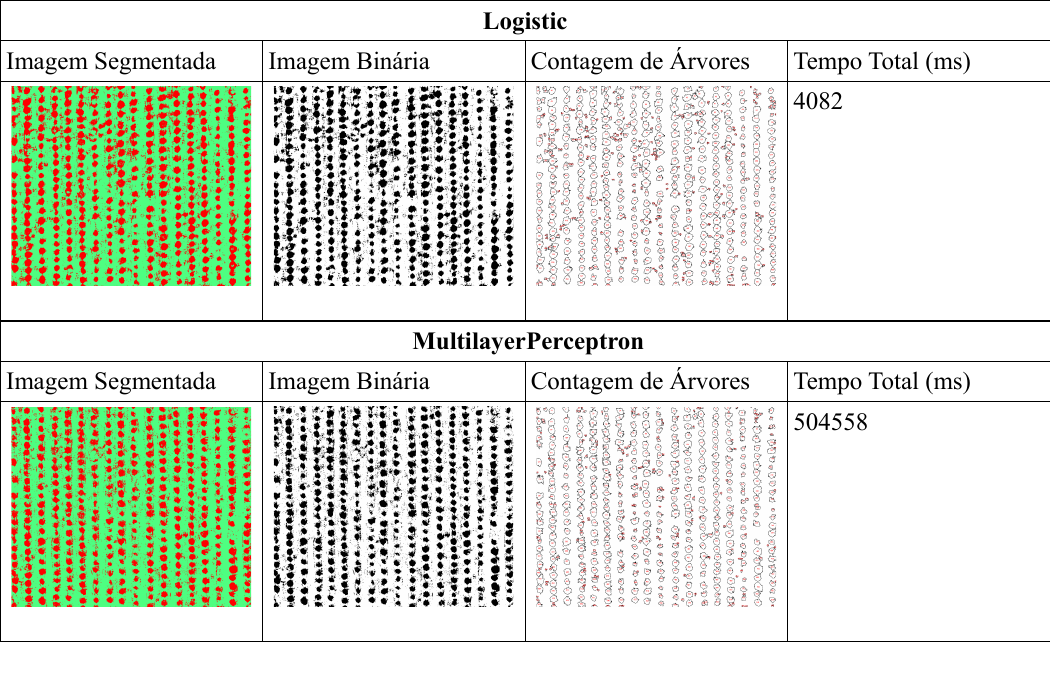}
\end{figure*}

\begin{figure*}[htbp!]
  \centering
    \includegraphics[width=1\textwidth,page=7]{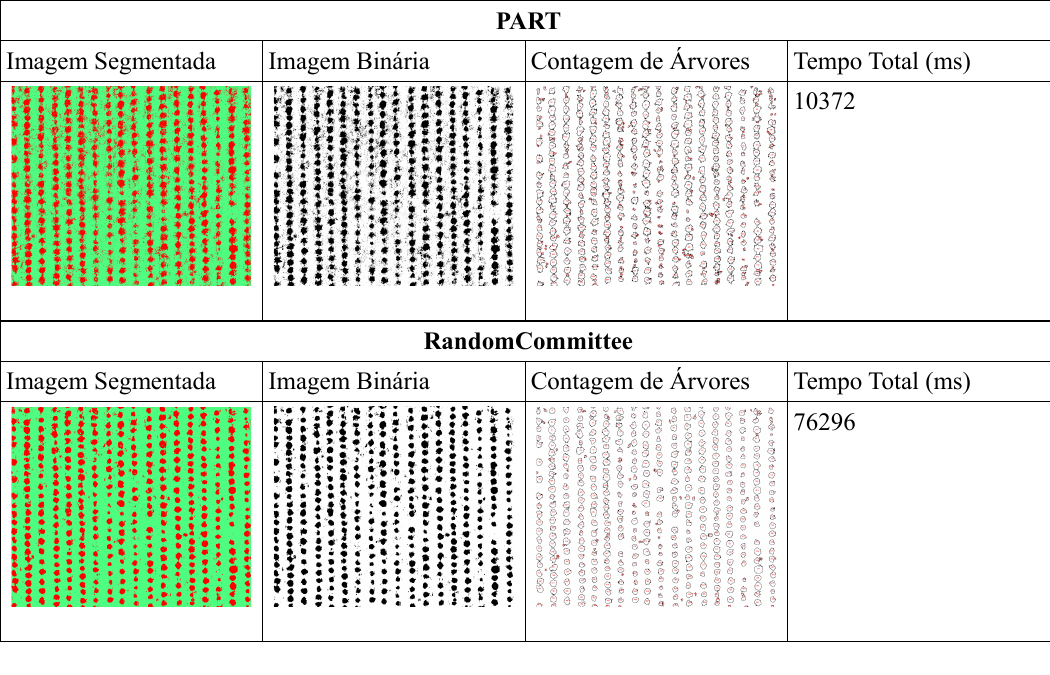}
\end{figure*}

\begin{figure*}[htbp!]
  \centering
    \includegraphics[width=1\textwidth,page=8]{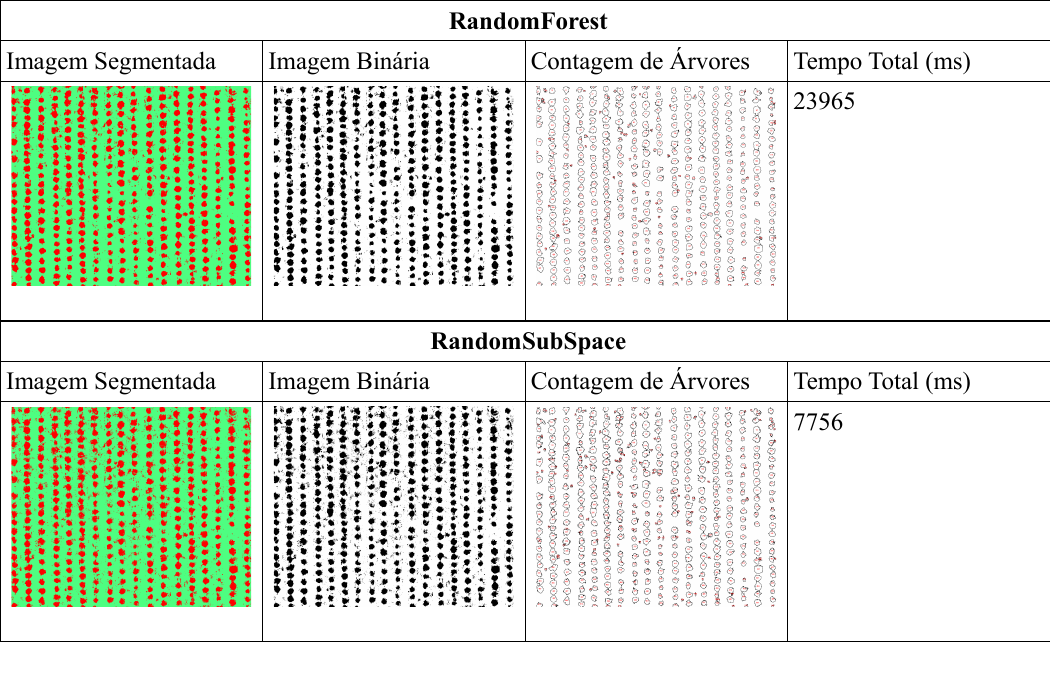}
\end{figure*}

\begin{figure*}[htbp!]
  \centering
    \includegraphics[width=1\textwidth,page=9]{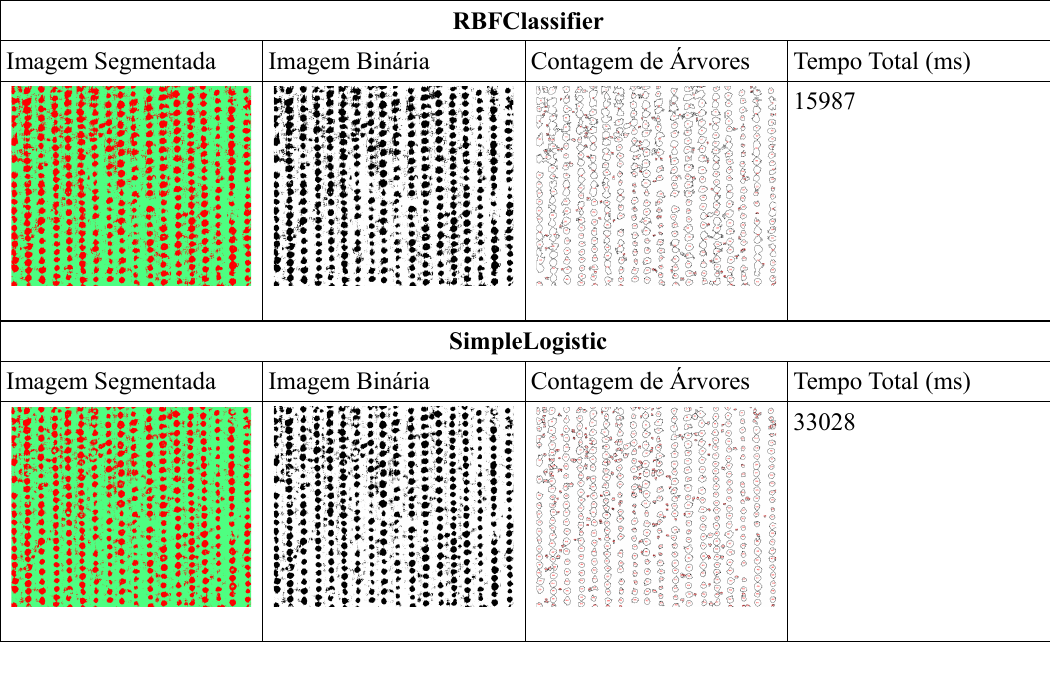}
\end{figure*}

\begin{figure*}[htbp!]
  \centering
    \includegraphics[width=1\textwidth,page=10]{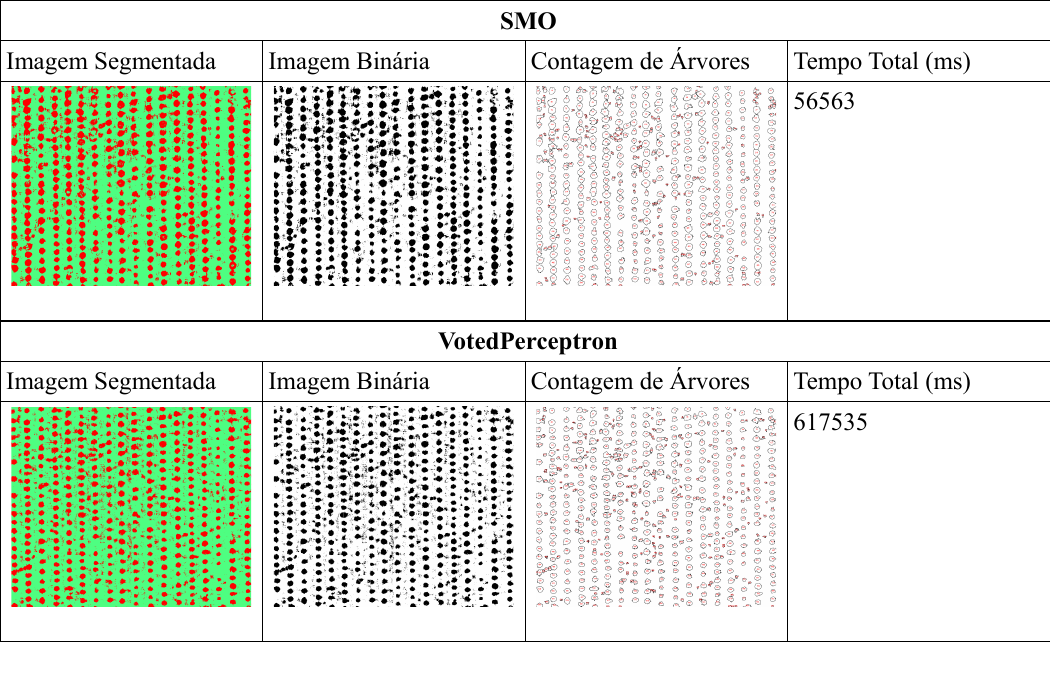}
\end{figure*}

\end{document}